\documentclass[lettersize,journal]{IEEEtran}
\usepackage{amsmath,amsfonts}
\usepackage{algorithmic}
\usepackage{algorithm}
\usepackage{array}
\usepackage{caption}
\usepackage{textcomp}
\usepackage{stfloats}
\usepackage{url}
\usepackage{verbatim}
\usepackage{graphicx}
\usepackage{cite}
\usepackage{threeparttable}
\usepackage{bbm}
\usepackage{color}
\usepackage{marvosym}
\usepackage{booktabs}
\usepackage{multirow} 
\usepackage{amsmath}
\usepackage[cmyk]{xcolor}
\newcommand{\js}[1]{{\color{black}#1}}

\begin{document}

\captionsetup[figure]{name={Figure},labelsep=period}

\title{Towards Grouping in Large Scenes with Occlusion-aware Spatio-temporal Transformers}

\author{Jinsong Zhang, Lingfeng Gu,
Yu-Kun Lai,~\IEEEmembership{Member,~IEEE}, Xueyang Wang, 
Kun Li\textsuperscript{\Letter},~\IEEEmembership{Member,~IEEE}

\thanks{This work was supported in part by the National Natural Science Foundation of China (62122058 and 62171317) and Science Fund for Distinguished Young Scholars of Tianjin (No. 22JCJQJC00040).}

\thanks{Jinsong Zhang, Lingfeng Gu and Kun Li are with the College of Intelligence and Computing, Tianjin University,
Tianjin 300350, China.}
\thanks{Yu-Kun Lai is with the School of Computer Science and Informatics, Cardiff University, Cardiff CF24 4AG, United Kingdom.
}
\thanks{Xueyang Wang is with the Department of Electronic Engineering, Tsinghua University, Beijing 100190, China.}
\thanks{Corresponding author: Kun Li (Email: lik@tju.edu.cn).}}
        % <-this % stops a space
%\thanks{This paper was produced by the IEEE Publication Technology Group. They are in Piscataway, NJ.}% <-this % stops a space
%\thanks{Manuscript received April 19, 2021; revised August 16, 2021.}}

\markboth{IEEE Transactions on Circuits and Systems for Video Technology}%
{Towards Grouping in Large Scenes with Occlusion-aware Spatio-temporal Transformers}


\maketitle

\begin{abstract}
Group detection, especially for large-scale scenes, has many potential applications for public safety and smart cities. Existing methods fail to cope with frequent occlusions in large-scale scenes with multiple people, and are difficult to effectively utilize spatio-temporal information.
In this paper, we propose an end-to-end framework, \textit{GroupTransformer}, for group detection in large-scale scenes. 
To deal with the frequent occlusions caused by multiple people, we design an occlusion encoder to detect and suppress severely occluded person crops. To explore the potential spatio-temporal relationship, we propose spatio-temporal transformers to simultaneously extract trajectory information and fuse inter-person features in a hierarchical manner. Experimental results on both large-scale and small-scale scenes demonstrate that our method achieves better performance compared with state-of-the-art methods. On large-scale scenes, our method significantly boosts the performance in terms of precision and F1 score by more than $10\%$. 
On small-scale scenes, our method still improves the performance of F1 score by more than $5\%$. The project page with code can be found at \url{http://cic.tju.edu.cn/faculty/likun/projects/GroupTrans}.
% \emph{We will release the code for research purposes.}
\end{abstract}


\section{Introduction}
\IEEEPARstart{G}{roup} detection is a fundamental task in computer vision that involves identifying groups of people from images or videos, which has numerous applications in human-centric analysis tasks such as abnormal detection \cite{zhang2014social,cocsar2016toward}, trajectory prediction~\cite{berenguer2020context, wang2021dynamic, 9025431, liu2020trajectorycnn} and group activity recognition~\cite{xing2021deep,5457461, 8769904, 1632036}.  In this work, our primary focus is on group detection in large-scale scenes, which has significant implications for public safety \cite{2015}, dynamic environments \cite{bain2019dynamic}, and smart cities~\cite{rios2015proxemics},~\cite{BENDALIBRAHAM2021100023}. 

\begin{figure}[h]
    \centering
    \includegraphics[width=1.0\linewidth]{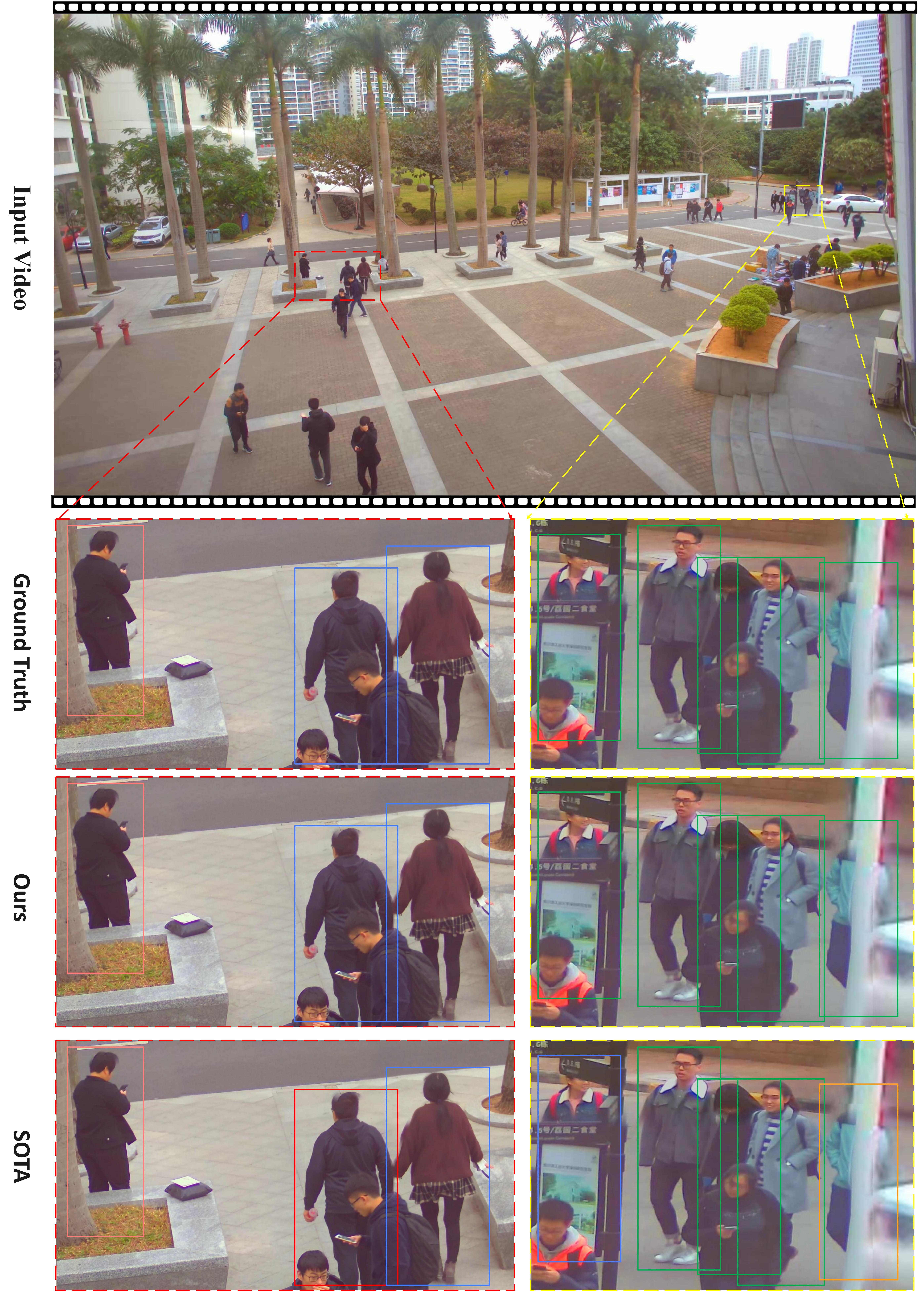}
    %\vspace{-0.2cm}
    \caption{Given a video of a large-scale multi-person scene, our method can detect reasonable groups, despite the large number of people and frequent occlusions in the scene. While the state-of-the-art method \cite{li2022self} fails to predict right results. Different groups are shown with different color boxes.}

    \label{result1}
\end{figure}

Existing group detection methods mainly focused on small-scale scenes with limited people or interactions. Some works focused on group detection on a single image. Traditional methods \cite{2015},\cite{10.1145/3462244.3479963},\cite{cristani2011social},\cite{8673233} utilized some hand-crafted features to identify F-formations \cite{kendon1990conducting} in crowds, which needed to detect the interaction spaces of groups of people. However, they only detected groups of F-formations, while other spatial patterns of groups were ignored. 
Furthermore, detecting groups based solely on a single image is not always reasonable, considering the dynamic nature of human relationships in real life. Consequently, researchers have explored group detection in videos. 
Some methods \cite{joint} directly input video clips into an I3D backbone \cite{8099985} to extract spatial and temporal information simultaneously.
However, it is important to note that spatial and temporal concepts are learned by different cognitive mechanisms in our brains \cite{Pang2020}, indicating the challenge of effectively learning both concepts within the same network.
Other methods \cite{ge2012vision}  have utilized trajectory information to detect groups in crowds. However, these approaches overlook the impact of image information, such as interactions between individuals, on group detection.
Moreover, the aforementioned existing methods are designed for group detection in short videos of small-scale scenes, which are not suitable for large-scale multi-person scenes due to the large computation cost.

Group detection on large-scale multi-person videos holds great research significance and offers a wide range of potential applications. However, there is a scarcity of works in this area primarily due to the limited availability of datasets designed specifically for large-scale multi-person scenes. Wang \textit{et al.}{}~\cite{wang2020panda} addressed this gap by introducing the gigaPixel-level humANcentric viDeo dAtaset (PANDA), which features a wide field-of-view, gigapixel-level image resolution, and temporally long-term crowd activities.  They also proposed a global-to-local zoom-in framework to detect groups in large-scale multi-person videos using the multi-modal inputs, \emph{i.e.}, images and trajectories. However, they ignored the frequent occlusions in crowd videos, \emph{i.e.}, a person can be occluded sometimes, and the appearance features at those time instances are  unreliable
for group detection. Furthermore, the method extracted image and trajectory features independently, employing 3D ConvNet \cite{8578783} for spatio-temporal image features and LSTM (Long Short-Term Memory) \cite{10.5555/2969239.2969329} for temporal trajectory features. It is worth noting that 3D ConvNet faces similar challenges as I3D, and LSTM fails to incorporate spatial information from trajectories. Recently, Li \textit{et al.}{}~\cite{li2022self} proposed a novel group detection method by pre-training the model using a self-supervised method, which produced promising results. 
However, this method also disregarded the occlusion problem and processed spatial and temporal information separately, which means that the spatial information and the temporal information are processed in sequence and not synchronized \cite{Pang2020}, limiting its performance.

In this paper, we propose GroupTransformer, an end-to-end spatio-temporal framework designed to address the challenges of frequent occlusions and complex spatio-temporal interactions in large-scale multi-person scenes. 
To tackle the issue of occlusions prevalent in large-scale scenes, we introduce an occlusion encoder that focuses on better extracting individual features. It leverages inter-frame similarity to identify and suppress features influenced by severe occlusions in specific frames, thereby enhancing the  robustness of our model to occlusions. The occlusion encoder improves the overall performance by effectively considering the inter-frame similarities of individuals.
Inspired by \cite{Pang2020}, we design a hierarchical scheme for the spatio-temporal transformers, which addresses the challenge of processing spatial and temporal information in a series computation manner. Our scheme incorporates a spatial branch and a temporal branch, enabling synchronized processing of spatial and temporal information during data-driven training. This design allows for comprehensive exploration of the relationship between trajectory and appearance, with high-level semantic features from the temporal branch guiding the learning of appearance features. The temporal information is primarily extracted from trajectory features using densely connected convolutional layers, while the spatial information is fused and obtained through a transformer encoder.
We evaluate our method on both small-scale scene datasets and large-scale scene datasets, and the experimental results demonstrate its superiority over state-of-the-art methods. Figure \ref{result1} illustrates an example of the grouping results compared our method with the state-of-the-art method, S3R2 \cite{li2022self}. To facilitate further research, we will release the code for GroupTransformer for academic purposes.

The main contributions of this work are summarized as follows:
\begin{itemize}
\item We propose an end-to-end framework which fuses spatio-temporal information from multi-modal inputs to detect groups in a large-scale multi-person scene, which effectively explores the multi-modal information and well deals with the occlusion problem. 

\item We propose spatio-temporal transformers, comprising a spatial branch and a temporal branch, which facilitate the hierarchical fusion of appearance and trajectory features.

\item To further enhance the personalized individual feature extraction process, we propose an occlusion encoder. 
This component effectively suppresses features affected by severe occlusions in specific frames, thereby improving the anti-occlusion ability of our model.

\item Experimental results on both large-scale and small-scale benchmark datasets demonstrate the superior performance of our proposed method.

\end{itemize}

We organize  the remainder of this paper as follows: in Section \ref{related}, we give a brief review of related work, including static methods, dynamic methods for group detection, occlusion-aware methods, and spatio-temporal methods. In Section \ref{method}, we introduce our proposed GroupTransformer, including the training strategy and loss functions.  In Section \ref{exp}, we first validate the effectiveness of our method through qualitative and quantitative experimental results, comparing it with several state-of-the-art methods. Subsequently, we conduct ablation studies to assess the impact of different components in our model. Additionally, we evaluate the robustness of our method by introducing noise to the bounding boxes and randomly reducing the bounding boxes of each person. Finally, we conclude and discuss our work in Section~\ref{con}.

\section{Related Work} 
\label{related}

In this section, we review  group detection on still images and videos. The methods can be classified into static methods and dynamic methods according to the input. Besides, we review the existing works about occlusion-aware methods and spatio-temporal methods to learn about the motivation of our model.

\subsection{Static   Methods}
Group detection in still images has been studied extensively in the literature. These methods aim to detect interactions and spatial arrangements among individuals in a single image.

Early static methods focused on detecting F-formations, which are spatial patterns formed by free-standing conversational individuals. Kendon \textit{et al.}{}\cite{kendon1990conducting} proposed the concept of F-formations and identified specific spatial configurations that tend to emerge during conversations. Traditional static methods for detecting F-formations relied on hand-crafted rules and mathematical models \cite{hung2011detecting,vascon2016detecting}. 
With the advancement of deep learning, recent static methods have benefited from the power of deep neural networks to improve performance in group detection. For example, some methods have adopted Graph Neural Networks (GNNs) \cite{47094} to model the mutual features and relationships among individuals in a single image. These methods leverage position and pose features of individuals and construct a fully connected graph to transfer messages and capture the group structure \cite{swofford2020improving}.

Static methods are suitable for closed scenes with a limited number of people, such as social gatherings or small group activities \cite{zen2010space,cristani2011social}. However, they have limitations in real-world scenarios where the number of individuals is large and dynamic. Additionally, static methods often rely on camera parameters and depth information to reconstruct world coordinates, which may not be readily available in practical settings.
Moreover, grouping results obtained from a single image may not be reliable, as the relationships and interactions between individuals are inherently dynamic and may change over time. Therefore, there is a need for methods that can leverage temporal information from video sequences to improve group detection performance.

\subsection{Dynamic Methods}
Dynamic methods leverage the temporal information from consecutive frames to infer the relationships among individuals in a video sequence. Early dynamic methods focused on using trajectory features for group detection. Ge \textit{et al.}{} \cite{ge2012vision} proposed a classic trajectory-based method that utilized pedestrian detection and tracking techniques to extract trajectories from video frames. They then applied hierarchical clustering to detect groups of people with similar trajectories. However, relying solely on trajectory information may not be sufficient to determine the relationships between individuals. For example, in a crowd walking on a sidewalk, most people may have similar trajectories but have no direct relationship with each other.

To overcome this limitation, appearance features obtained from images, which contain meaningful interactive information, are incorporated to obtain reliable grouping results. Ehsanpour \textit{et al.}{} \cite{joint} proposed a novel framework for small-scale videos group detection in small-scale videos by utilizing appearance features. They used the I3D network \cite{gu2018ava} as a video backbone to extract spatial and temporal features across frames. However, this type of feature extractor may not be suitable for large-scale scenes, especially for videos with gigapixel-level resolution. Additionally, using the same network to extract both spatial and temporal features from the inputs may not be reliable.

To address the challenges of large-scale scenes, researchers have explored different approaches.  
Wang \textit{et al.}{}~\cite{wang2020panda} proposed a global-to-local zoom-in framework for group detection in large-scale scenes. They utilized both appearance and trajectory features.  However, in their appearance-based model, the fusion of spatial and temporal information from appearance input was not well addressed. In their trajectory-based model, they used LSTM to capture temporal information from the trajectory input but ignored the spatial information. Furthermore, the occlusion of appearance features of different people in long-duration crowd videos was not properly considered, leading to inaccurate grouping results.
Recently, Li \textit{et al.}{}~\cite{li2022self} proposed a two-stage method that pre-trains the model on unsupervised tasks before fine-tuning for group detection.  While achieving promising performance, this approach neglects frequent occlusions in crowd videos and relies on a gated recurrent unit model for temporal information aggregation, which may limit its effectiveness. 

In this paper, we propose an end-to-end framework for group detection in large-scale multi-person scenes by extracting personalized individual features with an occlusion encoder and exploring the relationship between trajectory and appearance features using spatio-temporal transformers. By explicitly considering occlusion and leveraging both spatial and temporal information, our method aims to improve group detection performance in challenging large-scale scenes.

\subsection{Occlusion-aware Methods}

Occlusion problems exist in many computer vision tasks, including person re-identification \cite{wang2022feature}, optical-flow estimation \cite{wang2018occlusion}, instance segmentation \cite{ke2023occlusion} and so on.
Existing occlusion-aware methods aim to recover the occluded information through different approaches.
For person re-identification, Wang \textit{et al.}~\cite{wang2022feature} proposed a feature erasing and diffusion network to recover the occluded information by data augmentation to achieve intrinsic representation. 
For flow estimation, Wang \textit{et al.}~\cite{wang2018occlusion} addressed occlusion  by estimating an occlusion map to post-process the estimated flow. 
For instance segmentation, Ke \textit{et al.}~\cite{ke2023occlusion} presented a novel perspective that segments single images as double-layer images, and adopted a transformer-based network to recover the occluded information, which meant it also tries to deal with the occlusion problem by recovering the occluded information.

In our work, instead of recovering the occluded information, we adopt a transformer-like architecture \cite{vaswani2017attention} as our occlusion encoder to extract more useful and precise information,
while ignoring unreliable information caused by occlusions. By focusing on the essential information, our model aims to achieve accurate and reliable group detection in the presence of occlusions.

\subsection{Spatio-temporal Methods}
\label{discussst}
\js{The fusion of spatial and temporal information is a well-studied problem, particularly in video-based tasks \cite{https://doi.org/10.1049/cit2.12191, zheng2022abnormal, fang2022st, yang2019spatio, transvod, mlstformer, groupformer, cong2021spatial, 10057034, 10029908}. }Some methods used optical flow as the temporal information to cope with video instance segmentation \cite{tsai2016video, ding2020every}. However, these approaches heavily rely on accurate optical flow estimation, which can be challenging and memory-intensive, making them less suitable for large-scale scenes. 
\js{Some methods \cite{groupformer, mlstformer} utilize CNN-based networks to extract frame-wise features and employ RoIAlign to extract individual features for group activity recognition. However, in the context of large-scale scenes with images at the giga-pixel level, extracting frame-wise features becomes impractical. }Ehsanpour \emph{et al.} \cite{joint} employed video-based backbones, such as I3D \cite{8099985}, for video activity recognition. However, using the same architecture to extract both spatial and temporal features may not be optimal \cite{Pang2020}.
Zheng \textit{et al.}{}~\cite{zheng2022abnormal} proposed a novel temporal attention mechanism  for abnormal event detection. They treated temporal information as a sequence and used non-local attention to extract and aggregate it.  
Similar with \cite{zheng2022abnormal}, some methods \cite{liu2021decoupled, geng2022rstt, li2022learning} adopted transformer-like architectures to cope with spatio-temporal information. 
These methods consider the temporal information as a sequence, and apply attention mechanism to extract and aggregate information for each frame, which is suitable for video-based tasks. 
However, as discussed in \cite{Pang2020}, the spatial information and the temporal information should be learned by different cognitive mechanisms, and should be synchronized to process the sequential information.
\js{The conventional approaches \cite{10029908, 10057034, transvod} that first extract spatial information from each frame and then fuse it by temporal transformers may lead to a lack of information synchronization.}
Besides, in group detection, the temporal information \emph{e.g.}, trajectories, is not just a sequence, it is also the important location information for distinguishing the group results. 
LSGD \cite{wang2020panda} and S3R2 \cite{li2022self} handle temporal information and spatial information respectively, which can result in suboptimal performance.

Motivated by \cite{Pang2020}, we propose to update the temporal information and the spatial information in a hierarchical manner. Specifically, we propose the spatio-temporal transformer with a temporal branch and a spatial branch, which first extracts temporal information, and then concatenates with the spatial information. With this operation, the spatial information can be  synchronized with temporal information. We stack several spatio-temporal transformers to extract and aggregate spatial and temporal information effectively.

\vspace{4mm}
\section{Method}
\label{method}
\begin{figure*}[!t]
    \centering
    \includegraphics[width=0.89\linewidth]{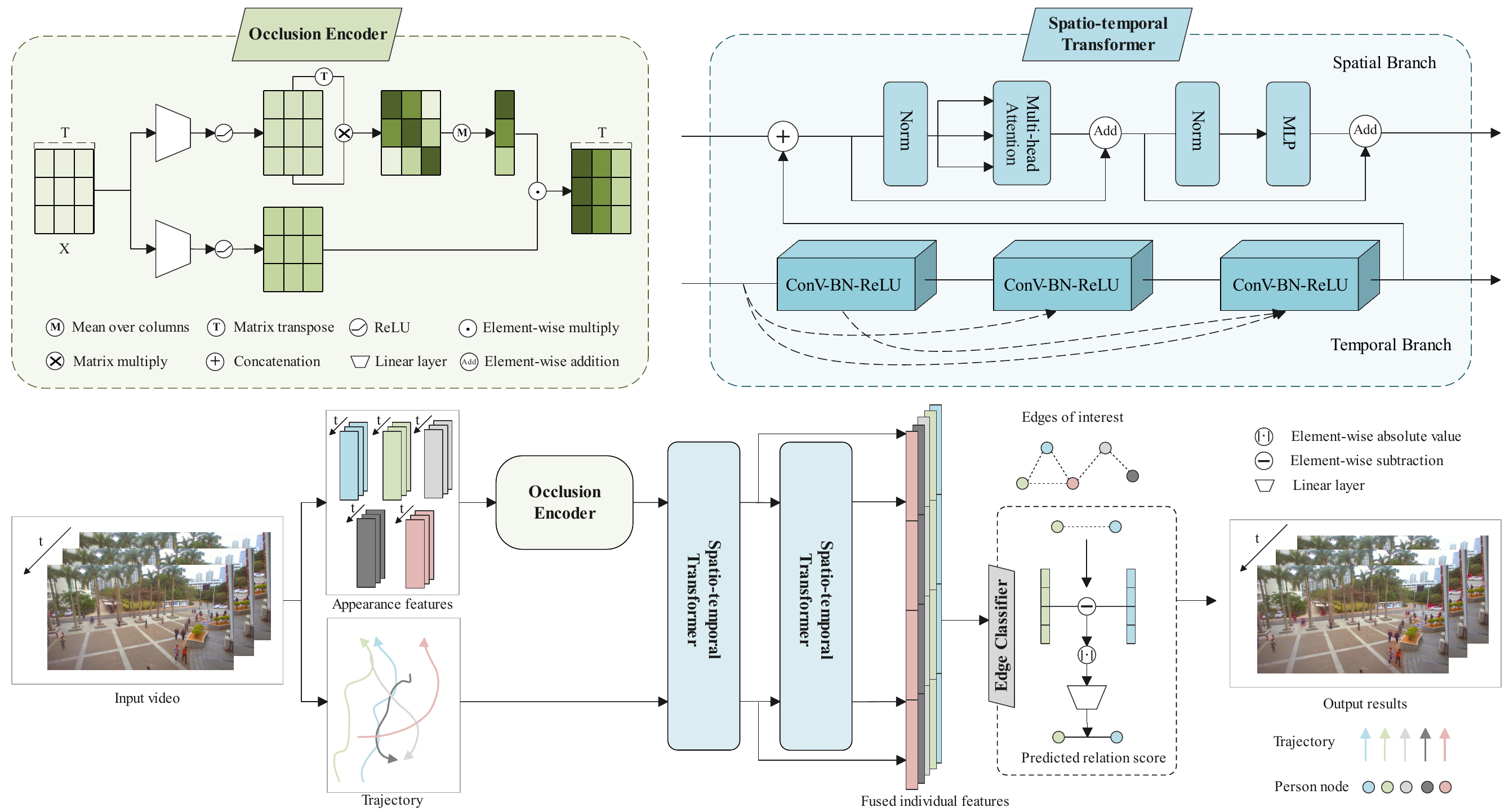}
    % \put(47,52){\small\bfseries  $\mathcal{Z}_{app_m}$}
    
    \caption{\js{Overview of the proposed GroupTransformer. We initially adopt a pre-trained ResNet50 network to extract individual appearance features. Subsequently, the proposed occlusion encoder is utilized to filter out occluded information. Finally, spatio-temporal transformers are employed to fuse both trajectory and appearance features. In output results, the individuals encapsulated within red bounding boxes, who are interconnected by blue edges, are classified into the same group.}}
    \label{framework}
\end{figure*}

The objective of our work is to detect groups in large-scale multi-person scenes. To achieve this, we represent individuals and their relationships as vertices and edges in a graph $G=(V,E)$. We then formulate the group detection task as an edge classification task, and exploit features from multiple modalities (the trajectory and the video) for edge classification. 
The proposed framework consists of three key components: 1) \textbf{occlusion encoder}, which takes the appearance feature of individuals as inputs and aims to alleviate the influence of occluded person crops in specific frames; 2) \textbf{spatio-temporal transformers (STT)}, which encode the features from both modalities, \emph{i.e.}, the trajectory and the video, and fuse information across both spatial and temporal dimensions; 3) \textbf{edge classifier}, which classifies the interested edges in the graph $G$ according to the fused individual features from spatio-temporal transformers. Figure \ref{framework} shows the framework of our method, depicting the flow of information and the interaction between the different components. 

\subsection{Occlusion Encoder}
Previous works \cite{wang2020panda, li2022self} of group detection in large-scale scenes ignored  the issue of frequent occlusions in multi-person scenes. As a result, the extracted features may contain noise and unreliable information. To address this problem, we propose the occlusion encoder, inspired by the attention mechanism \cite{vaswani2017attention}. Instead of recovering occluded information \cite{wang2022feature, wang2018occlusion, ke2023occlusion} in other vision tasks, the occlusion encoder aims to filter out the occluded information in the input sequence, allowing us to obtain more reasonable and meaningful information from the inputs.

The occlusion encoder consists of two learnable functions, denoted as $f(\cdot)$ and $g(\cdot)$. The input is the sequential appearance features $\mathcal{X}\in\mathbb{R}^{N\times D\times T}$ extracted by a pre-trained ResNet50 \cite{he2016deep}, where $N,D,T$ represent the number of people, the dimension of appearance features, and the number of frames, respectively. For each person and its sequential appearance features $X=[\mathbf{x}_{1},\mathbf{x}_{2},...,\mathbf{x}_T] \in\mathbb{R}^{D\times T}$, we assume that most of the person crops are not occluded. Then, we leverage the fact that appearance features for the same non-occluded person crops exhibit inter-frame similarity, while occluded ones are likely to have little similarity with other frames. 

To capture the inter-frame similarity, we calculate the affinity between frames using the inner product in the normalized feature space. Specifically, the similarity between frame $i$ and frame $j$ of a person can be computed as:
\begin{equation}
 s_{i,j}=\frac{f(\mathbf{x}_{i}) f(\mathbf{x}_{j})^\top}{\Vert f(\mathbf{x}_{i}) \Vert \Vert f(\mathbf{x}_{j}) \Vert} \label{eq:1}.
\end{equation}
Here, $f(\cdot)$ is a learnable function that maps appearance features to a feature space, and $\mathbf{x}_{i}$ represents the appearance feature 
at frame $i$. The similarity values range from 0 to 1, as $f(\cdot)$ outputs ReLU-activated feature vectors.

To determine the attention value $a_{i}$ for frame $i$, we calculate the mean similarity between frame $i$ and all other frames:
\begin{equation}
 a_{i}=\frac 1T\sum_{j=1}^{T} s_{i,j}\label{eq:2}.
\end{equation}
For a severely occluded frame $i$, it will have little similarity with the other frames, resulting in a small attention value $a_{i}$. 
In practice, if a person does not appear in all frames, the mean value is calculated only based on the visible frames. Finally, we obtain the processed appearance feature with the attention mask by:
\begin{equation}
 \mathbf{z}_{i}= g(\mathbf{x}_{i}) \times a_{i}\label{eq:3}.
\end{equation}
The embedding functions $f(\cdot)$ and $g(\cdot)$ are implemented by a linear layer and a ReLU activation function. The output of this module is the refined appearance features for each person, denoted as $Z_{app}=[\mathbf{z}_{1},\mathbf{z}_{2},...,\mathbf{z}_T]$.

\subsection{Spatio-temporal Transformers}
Previous works   on video understanding \cite{ge2012vision, joint} adopted RNN/LSTM-based  or 3D ConvNet-based  approaches to extract temporal information. However, these methods typically process spatial and temporal information in a sequential manner, leading to a lack of synchronization between spatial and temporal features \cite{Pang2020}. This limitation hinders the effective representation of extracted features. In order to exploit the potential relationship between the multi-modal inputs in both the spatial and temporal dimensions, we propose the spatio-temporal transformers (STT) to fuse spatio-temporal information from trajectory and appearance features.

The STT module contains two branches: a temporal branch and a spatial branch. The temporal branch extracts  high-level temporal semantics from trajectory  features, such as the moving velocity at each moment. We design the temporal branch based on the DenseNet \cite{huang2017densely}, where 1D convolutional layers are utilized, and each layer is connected to every other layer in a feed-forward manner. This architecture enables the preservation of both low and high-level trajectory semantics.

The spatial branch  employs a transformer encoder to capture the spatial patterns from both trajectory and appearance features. 
Specifically, given the input appearance feature of all people $\mathcal{Z}_{app_{m}} = \{Z_{app_m^{n}}|n=1,2,...,N\}$  at depth $m$, it is firstly concatenated with the processed trajectory features $\mathcal{Z}_{traj_{m+1}}$ to form the raw embedding features for individual people:
\begin{equation}
 \mathcal{Z}_{m}= \mathcal{Z}_{app_{m}} \oplus \mathcal{Z}_{traj_{m+1}}\label{eq:4}.
\end{equation}

We view the temporal dimension as the batch dimension and apply a transformer encoder to exploit spatial context for all people. The process of embedding spatial context for frame $i$ can be formulated as:
\begin{equation}
 Q_{m}^{i}=  \mathcal{Z}_{m}^{i}W_{q,m}, K_{m}^{i}=  \mathcal{Z}_{m}^{i}W_{k,m},V_{m}^{i}=  \mathcal{Z}_{m}^{i}W_{v,m} \label{eq:5},
\end{equation}
\begin{equation}
  V_{m}'^{i}=softmax(\frac{Q_{m}^{i}K_{m}^{i}}{\sqrt{D_1}})V_{m}^{i}+V_{m}^{i}\label{eq:6},
\end{equation}
\begin{equation}
  V_{m}''^{i}=\mathrm{MLP}(V_{m}'^{i})\label{eq:7},
\end{equation}
where $W_{q,m}$,$W_{k,m}$,$W_{v,m}$ are learnable parameters, ${D_1}$ is the dimension of ${Q}_{m}$,
and $\mathrm{MLP}$ is the Multi-Layer Perceptron in the canonical transformer. The features of all people at all time instances $\{V_m''^{i}|i=1,2,...,T\}$ are packed together as $\mathcal{Z}_{app_{m+1}}$. 
Finally, the STT module outputs the extracted features from the two branches,  $\mathcal{Z}_{traj_{m+1}}$ and $Z_{app_{m+1}}$, which can be further used as  input of the next STT module. We stack $M$ STT modules to form a deep model inspired by \cite{zhang2020human}.

\begin{table*}[htbp]
  \centering
  \caption{Architecture of GroupTransformer. The hyper-parameters for linear layers are denoted by $L$ (input dimension, output dimension); the hyper-parameters for convolutional layers are denoted by $C$ (input dimension, output dimension, kernel size); and the hyper-parameters for transformer encoders are denoted by $E$ (hidden layer dimension, hidden layer number, the number of heads). The variables $N,T,N_e$ represent the number of people, the number of frames, and the number of edges, respectively.}
  \setlength{\tabcolsep}{6mm}{
    \begin{tabular}{c|c|c|c}
    \toprule
    Module & Input & Parameter & Value \\
    \hline
    \multirow{2}[2]{*}{Occlusion Encoder} & \multirow{2}[2]{*}{$N\times T\times 16384$} & $f(x)$  & $L(16384, 1024)$ \\
          &       & $g(x)$  & $L(16384, 512)$ \\
    \hline
    \multirow{4}[2]{*}{STT1} & $N\times 5\times T$ & Conv1 & $C(5,64,3)$ \\
          & $N\times 69\times T$ & Conv2 & $C(69,64,3)$ \\
          & $N\times 133\times T$ & Conv3 & $C(133,128,3)$ \\
          & $T\times N\times 640$ & Transformer Encoder & $E(128,2,4)$ \\
    \hline
    \multirow{4}[2]{*}{STT2} & $N\times 128\times T$ & Conv1 & $C(128,64,3)$ \\
          & $N\times 192\times T$ & Conv2 & $C(192,64,3)$ \\
          & $N\times 256\times T$ & Conv3 & $C(256,128,3)$ \\
          & $T\times N\times 256$ & Transformer Encoder & $E(128,2,4)$ \\
    \hline
    Edge Classifier & $N_e\times T\times 512$ & Linear & $L(512,1)$ \\
    \bottomrule
    \end{tabular}%
}
  \label{fm}%
\end{table*}%

\subsection{Edge Classifier}
\js{Previous spatio-temporal transformers \cite{mlstformer, groupformer} designed for group activity recognition are not suitable for grouping. Adopting these approaches to generate the complete output results simultaneously makes it necessary to predict all pairwise relation scores. This way not only imposes additional computational overhead but also introduces training complexities owing to the abundance of zero scores, given that most individuals are not part of the same group. To deal with this problem, }we propose an edge classifier to predict the relation scores for edges in the graph $G(V,E)$ using the features of individual people. The individual features are collected by concatenating all appearance and trajectory features from the STT modules of different depths, which can be expressed  as:
\begin{equation}
 \mathcal{Z}_{all}= \mathcal{Z}_{app_{1}} \oplus ... \oplus \mathcal{Z}_{app_{M}} \oplus \mathcal{Z}_{traj_{1}} \oplus ... \oplus \mathcal{Z}_{traj_{M}}. \label{eq:8}
\end{equation}
Note that $\mathcal{Z}_{all}$ preserves the temporal dimension, resulting in $\mathcal{Z}_{all}=[\mathbf{z}^{1}_{all},\mathbf{z}^{2}_{all},...\mathbf{z}^{T}_{all}]$. Then, we convert the individual features to inter-person edge features. Specifically, for an edge 
$(u,v)\in E$, we construct its feature $F_{(u,v)}$ by  taking the absolute difference between the features of the two individuals:
\begin{equation}
 {F}_{(u,v)}=\left|\mathcal{Z}^u_{all} - \mathcal{Z}^v_{all}\right|. \label{eq:9}
\end{equation}

After obtaining the edge features, they are fed into a Multi-Layer Perceptron (MLP) to generate frame-wise classification logits. Subsequently, a global average pooling is applied along the temporal dimension to obtain the overall prediction. This can be expressed as:

\begin{equation}
 R_{u,v} = \mathrm{MLP_R}(F_{u,v}),~ c_{u,v} = \frac 1T\sum_{t=1}^T {R}_{u,v}^{t}, \label{eq:10}
\end{equation}
where $R_{u,v}$ represents the classification logits for edge $u$ and edge $v$, $c_{u,v}$ represents the average prediction score over the temporal dimension, and $\mathrm{MLP_R}$ is the Multi-Layer Perceptron layer.
For the person pairs with social interactions, the classifier will assign positive scores for the corresponding edges, while for those without interactions, negative scores are supposed.

\subsection{Training}

We train the proposed model in an end-to-end manner. Instead of sampling a fixed number of people in a scene, we sample a fixed number of groups, which ensures a certain number of positive edges to guarantee sufficient training. 
However, the number of negative edges is still much higher than that of positive edges. This leads to two issues: (1) classifying a large number of edges becomes computationally inefficient; and (2) most of the negative edges are too easy to discriminate,  limiting the performance of the model. 

To cope with these problems, we propose a pre-processing strategy that filters out edges that are not worth training during the construction of the relation graph. We keep all positive edges due to their rarity, while removing unnecessary negative edges. Specifically, we first filter out edges where the two people never appear simultaneously in the frames. Then, we ignore person pairs with a minimal distance on trajectories larger than a threshold $\delta_{train}$. This means that only hard negative edges are retained for training. 
By applying this strategy, we construct the edge set $E$ corresponding to the sampled people. Simultaneously, the edges are assigned binary labels $y_{u,v}\in {0,1}, \forall (u,v)\in E$ according to the ground-truth group annotation.
During training, we utilize binary cross-entropy loss to train the model in a supervised manner:

\begin{align}
 \mathcal{L}=-\sum_{(u,v)\in E} (1-\lambda)y_{u,v}\times \log(\sigma (c_{u,v}))+ \notag\\
 \lambda(1-y_{u,v})\times \log(1- \sigma(c_{u,v})),
\end{align}
where $\sigma(\cdot)$ denotes the sigmoid function, and $\lambda = \frac{|E_{positive}|}{|E|}$ is a balance coefficient to account for the difference in the number of positive and negative edges. Here, $E_{positive}$ and $E$ are the sets positive edges and all edges, and $|\cdot|$ is the cardinality of the set.

\subsection{Inference}
During the training phase, the input people are controlled to form a limited number of groups. However, during inference, all the people from a scene are fed into the model simultaneously. In the case of large scenes, this can result in thousands of people and millions of edges in the graph, which is computationally expensive. To improve efficiency, we adopt two strategies  to filter out obvious negative edges during classification.

The first strategy is similar to the training phase. We filter out edges where the distance between two people exceeds a threshold $\delta_{test}$. It is important to note that $\delta_{test}$ should be larger than $\delta_{train}$ to avoid mistakenly removing positive edges. The second strategy involves removing edges where two people do not appear simultaneously for most of the frames. We calculate the Intersection over Unions ($IoU$) at visible frames of each person. If the $IoU$ is less than a threshold $\gamma$, the edge is assumed to represent no interaction behavior and is removed from the edge set $E$.
After applying these filtering strategies, we feed the remaining edges into our model and construct an affinity matrix using the predicted relation scores. The ignored edges are assigned a score of 0 by default. Finally, we solve a clustering problem on the affinity matrix to detect groups. Various clustering algorithms can be applied for this task (see the next subsection for implementation details).

\subsection{Implementation Details}
The appearance features are extracted using a ResNet50 \cite{he2016deep} pre-trained on ImageNet. 
\js{We extract feature vectors of dimension 4$\times$2$\times$2048 from the layer preceding the max-pooling operation in ResNet50. These vectors are then flattened to form 16384-dimensional feature vectors, which serve as the appearance features.} Our models are trained using a stochastic gradient descent optimizer with no momentum, and the learning rate is set to 0.1 initially. We sample 8 groups in each iteration and apply the gradient descent every 10 iterations. For large-scale datasets, the models are trained for 200 epochs, and the learning rate drops by a factor of 5 at 50, 100 and 150 epochs. For small-scale datasets, the models are trained for 20 epochs without learning rate decay. 
In  data preprocessing, we set $\delta_{train} = 0.1, \delta_{test} = 0.2, \gamma = 0.3$ for large-scale scenes, and $\delta_{train} = 0.5, \delta_{test} = 0.75, \gamma = 0.001$ for small-scale scenes. It is important to note that the positions in trajectories are normalized, so larger scenes with high image resolution require smaller threshold values. For group inference, we apply the same clustering algorithm as the compared methods to ensure fair comparison. We use  label propagation \cite{8764496} for large-scale scenes and spectral clustering \cite{ng2002spectral} for small scenes.
We train and test our model on a desktop with an Intel(R) Xeon(R) CPU and a GeForce RTX 2080 Ti GPU. 
The training time for PANDA dataset is about 20 hours, while it takes around 3 hours for JRDB dataset.
\js{The inference time of our model is related to the number of edges in the test video. On average, the inference time on JRDB test set is 0.11 seconds, while the inference time on PANDA test set is 0.48 seconds.}
The number of model parameters is about 33.73M.

\textbf{Detailed Framework.}
The detailed architecture of our GroupTransformer is defined in Table \ref{fm}. We stack 2 Spatio-Temporal Transformers (STT), namely STT1 and STT2. The $f(x)$ and $g(x)$ functions are followed by a ReLU activation function. All the Conv1, Conv2, and Conv3 are followed by a Batch Normalization layer \cite{Chollet_2017_CVPR} and a ReLU activation function.

\begin{table}[htbp]
  \centering
  \caption{Quantitative comparison on \textit{PANDA} dataset.}
    \scalebox{1.0}{\begin{tabular}{c|ccc}
    \hline
   \raisebox{-1.4 ex}[0cm][0cm]{Method} & \multicolumn{3}{c}{PANDA}\\
       &     Precision & Recall & F1  \\

    \hline
    
    G2L w/o Local  \cite{wang2020panda} & 0.237 & 0.120  & 0.160\\
    G2L w/ Random \cite{wang2020panda} & 0.244 & 0.133 & 0.172  \\
    G2L w/ Uncertainty \cite{wang2020panda} & 0.293 & 0.16 & 0.207 \\
    \hline
    Dis.Mat+ \cite{zhan2018consensus} & 0.429 & 0.120 & 0.188\\
    GNN w/ GRU & 0.419 & 0.173 &0.245  \\
    ARG \cite{wu2019learning} & 0.349 & 0.200 & 0.254 \\
    S3R2 \cite{li2022self} & 0.559 & 0.507 & 0.532 \\
    
    \hline
          Ours  & \textbf{0.750}  & \textbf{0.545} & \textbf{0.632}  \\
    
    \hline
    \end{tabular}%
    }
  
  \label{cmpPANDA}%
\end{table}%

\begin{table}[htbp]
  \centering
  \caption{Quantitative comparison on \textit{JRDB-Group} dataset.}
   \scalebox{1.0}{\begin{tabular}{c|ccc}
    \hline
   \raisebox{-1.8 ex}[0cm][0cm]{Method} &  \multicolumn{3}{c}{JRDB-Group}\\
       &     Precision & Recall & F1 \\

    \hline
    
    Joint \cite{joint}  &  0.300 & 0.284 & 0.291 \\
    JRDB-Group \cite{jrdb-group} & 0.390 & 0.379 & 0.384 \\
    \hline
    Dis.Mat+ \cite{zhan2018consensus}  & 0.573 & 0.235 & 0.334\\
    GNN w/ GRU  & 0.434 & 0.286 & 0.345 \\
    ARG \cite{wu2019learning} &  0.325 & 0.384 & 0.352\\
    S3R2 \cite{li2022self} &  0.577 & 0.562 & 0.569\\
    
    \hline
          Ours  &  \textbf{0.662} & \textbf{0.606} & \textbf{0.633} \\
    
    \hline
    \end{tabular}%
   }
  \label{cmpJRDB}%
\end{table}%

\begin{figure}[htbp]
    \centering
    \includegraphics[width=1\linewidth]{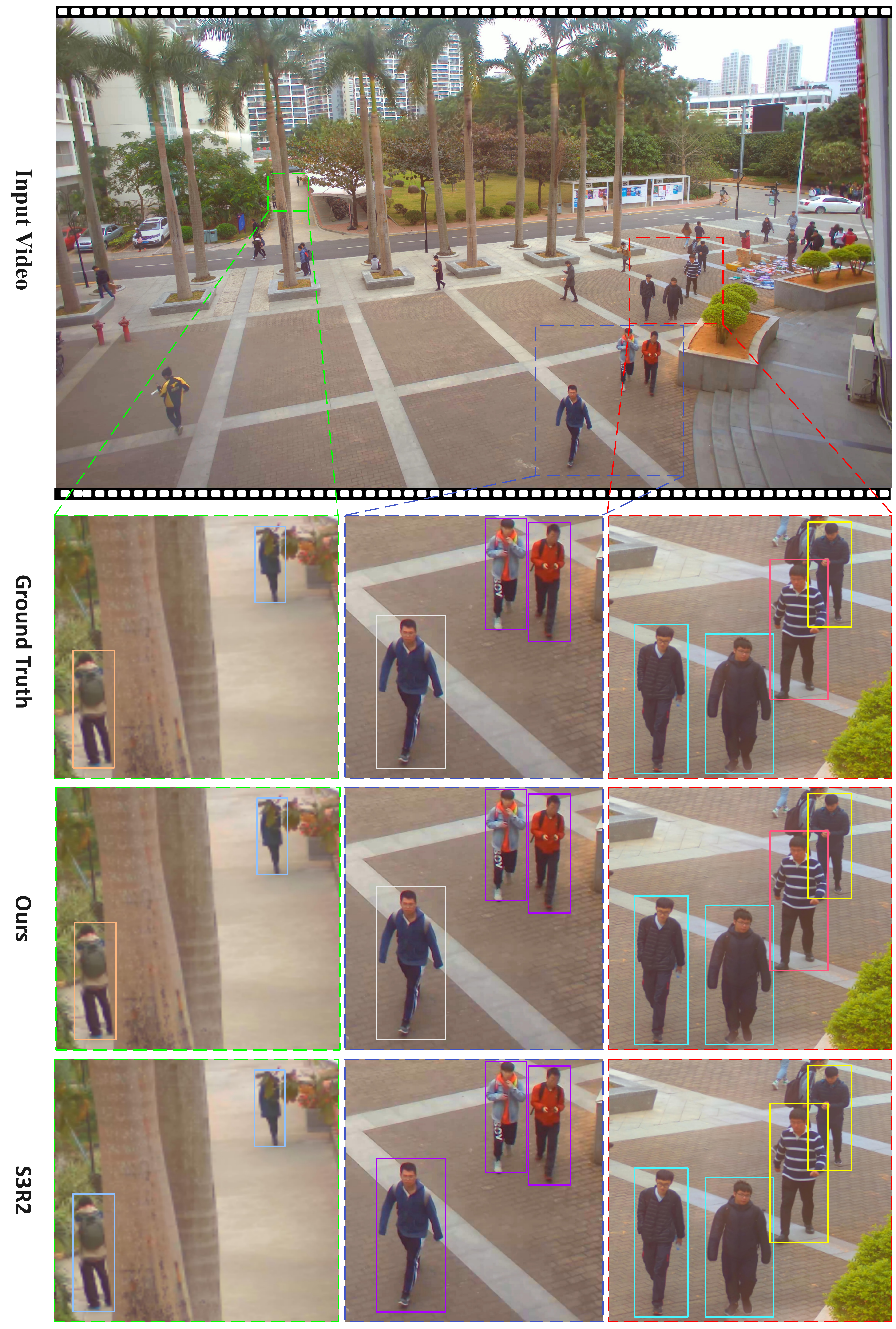}
 
    \caption{\js{Qualitative results compared with S3R2 \cite{li2022self} on \textit{PANDA} benchmark. More results can be found in the supplementary video.}}
  
    \label{companda}
\end{figure}

\begin{figure*}[htbp]
    \centering
    \includegraphics[width=0.95\linewidth]{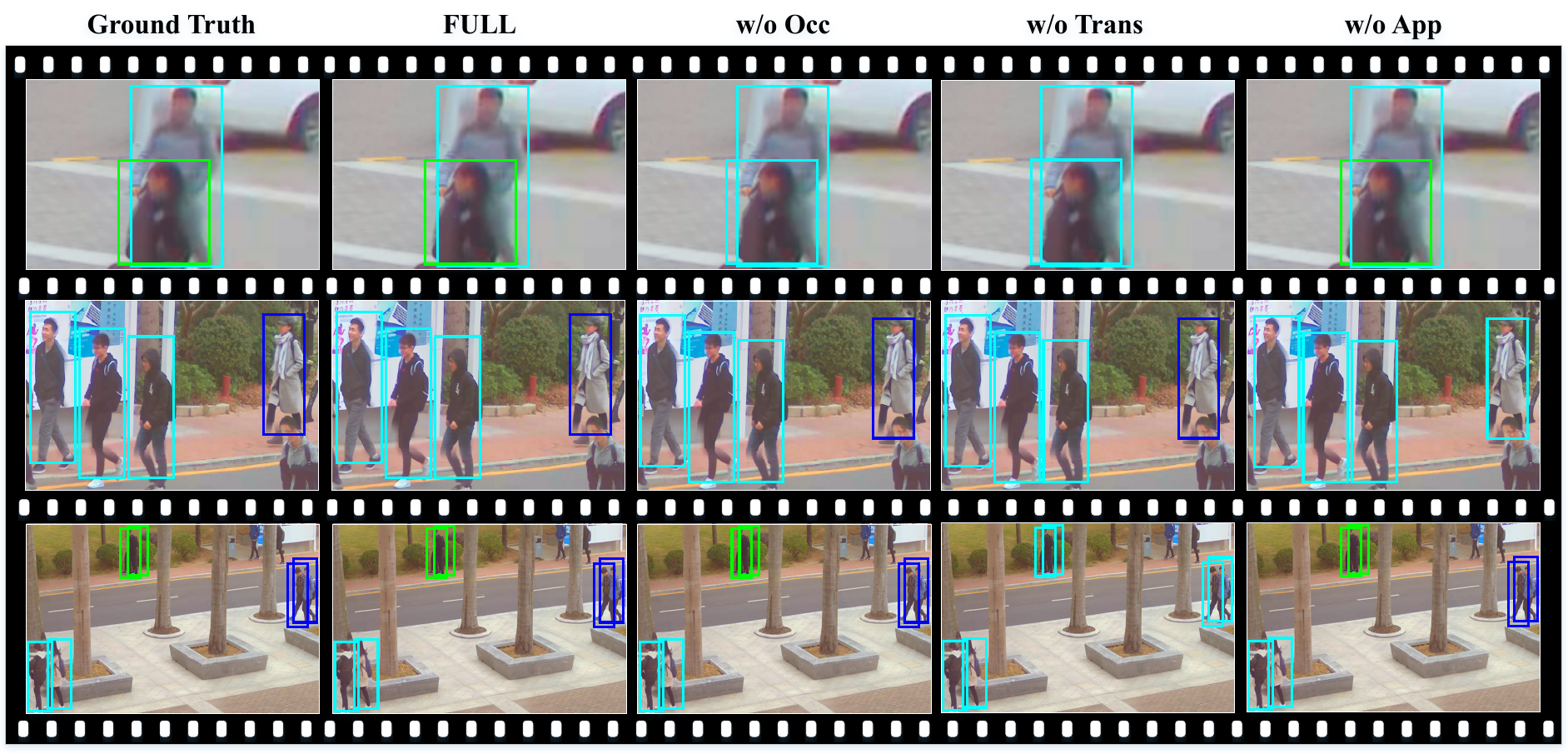}
  
    \caption{Qualitative results of ablation study on \textit{PANDA} benchmark.}

    \label{ablation}
\end{figure*}

\section{Experimental Results}
\label{exp}
\subsection{Datasets}
To validate the effectiveness of our method, we conduct our experiments on a large-scale dataset \textit{PANDA} \cite{wang2020panda} and a small-scale dataset \textit{JRDB-Group} \cite{jrdb-group} following \cite{li2022self}. The details of two datasets are given in the following.

\noindent
\textbf{PANDA benchmark.}
PANDA is a gigapixel-level human-centric video dataset with a wide field-of-view (up to $1 km^2$) and a very dense crowd (up to 4k subjects in a frame). It consists of 9 videos for group detection, providing rich and hierarchical ground-truth annotations, including bounding boxes, fine-grained labels, trajectories, and interactions. Following \cite{wang2020panda, li2022self}, we adopt 8 videos as the training set and 1 video as the test set for fair comparison. On average, each training video  has 2713 frames and 1070.4$k$ bounding boxes, while each test video has 3500 frames and 335.2$k$ bounding boxes. In the training and test video sets, the average group sizes per video are 144.6 and 75, respectively. Same with \cite{wang2020panda, li2022self}, we use the ground-truth bounding boxes and trajectories to validate the effectiveness of our method in training and test phases.

\noindent
\textbf{JRDB-Group benchmark.}
JRDB-Group  \cite{jrdb-group} is a multi-person video dataset captured by a panoramic camera equipped on a robot, which contains outdoor and indoor crowd scenes.  Following \cite{li2022self}, we take 20 videos as the training set and 7 videos as the test set, and use key frames to validate the performance of group detection. The key frames are sampled every 15 frames, and we get 1419 samples for training and 404 samples for testing.

\subsection{Metrics}
To evaluate the performance of our method, we use the same metrics as the compared state-of-the-art methods \cite{wang2020panda, li2022self}. 
For large-scale scenes, we calculate precision, recall, and F1-score using the half metric \cite{43104} with a group member IoU threshold of 0.5. The half metric determines whether a detected group is considered positive or negative based on the intersection over union (IoU) between the individuals in the detected group and the ground-truth group.
This can be formulated as
\begin{equation}
\frac{ \left|Gp_{det}\cap Gp_{gt}\right|}{max(\left|Gp_{det}\right|,\left|Gp_{gt}\right|)}>0.5,
\end{equation}
where $Gp_{det}$ denotes the detected groups, and $Gp_{gt}$ denotes the ground-truth groups.
That means if the number of individuals in the intersection is greater than half the number of individuals in the ground-truth group, the detected group is considered a positive sample. Precision, recall, and F1-score are then calculated based on these positive and negative samples.

\subsection{Comparison Results}

We try to conduct as much comparative experiments as possible to demonstrate the performance of our model. However, the source code of most group detection methods are not available, so we implement all models based on  \cite{li2022self} to give a clear comparison results. Here, we give a brief introduction of compared methods.

\noindent
\textbf{Dis.Mat+ \cite{zhan2018consensus}.} This method is a straightforward method that first measures the distance of each subject pair in the scene and adopts label propagation algorithm \cite{li2022self} to obtain the final group results. 

\noindent 
\textbf{GNN w/ GRU. }This method is designed by \cite{li2022self}, which  uses a gated recurrent unit model  to aggregate the temporal information and adopts a graph neural network to model the graph relation. 

\noindent 
\textbf{ARG \cite{wu2019learning}.} This method is a state-of-the-art method for group activity recognition. We follow the modifications applied in \cite{li2022self}. 

\noindent
\textbf{G2L \cite{wang2020panda}.} This method is the baseline method proposed with the PANDA benchmark. It models the group relation of subjects as a graph, and adopts a global-to-local strategy to leverage the visual cues for group detection. The group results can be obtained after label propagation. 

\noindent
\textbf{Joint \cite{joint} and JRDB-Group \cite{jrdb-group}.} The Joint method leverages group detection results to get better results for group activity recognition, which is suitable to be a compared method. The JRDB-Group is built on Joint by introducing spatial information.

\noindent
\textbf{S3R2 \cite{li2022self}. } This method is the most relevant method and is also the state-of-the-art method of group detection. It first trains the model using its self-supervised method, and then adds a group detection head to tune the model to get the group detection result using supervised training.

The quantitative results presented in Table \ref{cmpPANDA} and Table \ref{cmpJRDB} demonstrate the superior performance of our method compared to other state-of-the-art methods on the PANDA and JRDB-Group datasets.

On the PANDA dataset, our method achieves the best performance in terms of precision, recall and F1-score. The baseline method G2L and other straightforward methods struggle to predict correct group detection results for the large-scale dataset, resulting in unsatisfactory performance. S3R2, which adopts self-supervised learning for pre-training, performs better than the baseline methods but fails to consider occlusion circumstances in crowd scenes. Our method addresses the occlusion problem through the proposed occlusion encoder and leverages the spatio-temporal transformer to model temporal information. As a result, our method significantly improves the performance compared to S3R2, with an increase in the F1-score from 0.532 to 0.632. These results highlight the effectiveness of our approach in handling large-scale group detection challenges. 
Similarly, on the JRDB-Group dataset, our method outperforms the compared methods, including Joint and JRDB-Group, as well as Dis.Mat+, GNN w/ GRU, ARG, and S3R2. Our method achieves a higher precision, recall, and F1-score, indicating its superiority in group detection on both large-scale and small-scale scenes. Overall, the comparison results demonstrate that our method surpasses other methods in terms of performance metrics on both datasets, proving its effectiveness and suitability for group detection tasks.

Figure \ref{companda} presents qualitative results on the PANDA benchmark, comparing our approach with the state-of-the-art method S3R2 \cite{li2022self}\footnote{The public repository only contains the source code for the PANDA benchmark.}. It can be observed that S3R2 fails to detect groups in cases where individuals are walking towards the same direction, such as the three pedestrians  in the second column. With our spatio-temporal transformer, our model can distinguish whether pedestrians with similar trajectories belong to the same group. Dynamic results and additional results can be found in the supplementary video.

\begin{table*}[htbp]
  \centering
  \caption{Quantitative comparison with three alternative designs.}
  \setlength{\tabcolsep}{5.5mm}{
   \scalebox{1.0}{ \begin{tabular}{c|ccc|ccc}
    \hline
    \multicolumn{1}{c|}{\raisebox{-1.4 ex}[0cm][0cm]{Method}} & \multicolumn{3}{c|}{\textit{PANDA}} & \multicolumn{3}{c}{\textit{JRDB-Group}}    \\
\cline{2-7}          & \multicolumn{1}{c}{Precision} & \multicolumn{1}{c}{Recall} & \multicolumn{1}{c|}{F1} & \multicolumn{1}{c}{Precision} & \multicolumn{1}{c}{Recall} & \multicolumn{1}{c}{F1}  \\
    \hline
    w/o Occ. &0.686       &0.530       &0.598   &0.568      &0.522       &0.544      \\
    w/o Trans.  &0.574       &0.470       &0.517   &0.560       &0.508      &0.533  
    \\
    w/o App. &0.679       &0.545       &0.605   &0.585       &0.570       &0.577  \\
    FULL  &\textbf{0.750}      &\textbf{0.545}       &\textbf{0.632}  &\textbf{0.662} & \textbf{0.606} & \textbf{0.633}  \\
    \hline
    \end{tabular}%
%\vspace{-0.2cm}
}}
 
  \label{tab:abalation}%
\end{table*}%

\begin{table}[!htbp]
  \centering
  \caption{Quantitative results with different Gaussian noises.}
  \label{gs}%
  %\vspace{-0.3cm}
    \scalebox{1.0}{\begin{tabular}{c|ccc}
    \hline
    Metrics & \multicolumn{1}{c}{$\sigma=0$} & \multicolumn{1}{c}{$\sigma=0.1$} & \multicolumn{1}{c}{$\sigma=0.3$} \\
    \hline
    precision &0.750 & 0.750  & 0.715\\
    recall &0.545 & 0.545 & 0.409 \\
    F1    &0.632 & 0.623 & 0.519 \\
    \hline
    \end{tabular}%
  }
\end{table}%

\begin{table}[!htbp]
  \centering
  \caption{Quantitative results with different Missing detection rate.}
  \label{mdr}%
  %\vspace{-0.3cm}
    \scalebox{1.0}{\begin{tabular}{c|ccc}
    \hline
    Metrics & \multicolumn{1}{c}{\emph{MDR}=0} & \multicolumn{1}{c}{\emph{MDR}=0.1} & \multicolumn{1}{c}{\emph{MDR}=0.2} \\
    \hline
    precision &0.750 & 0.793  & 0.700\\
    recall &0.545 & 0.348 & 0.212 \\
    F1    &0.632 & 0.484 & 0.326 \\
    \hline
    \end{tabular}%
}
\end{table}%

\subsection{Ablation Study}

We evaluate our method with three alternative models to assess the factors that contribute to achieving better group detection results on both small-scale videos and large-scale videos.

\noindent
\textbf{The Model without Occlusion Encoder (w/o Occ.)}. We delete the mask branch in occlusion encoder and the appearance features are simply processed with a linear projection layer.

\noindent\textbf{The Model without Transformer (w/o Trans.)}. We use a MLP to fuse appearance and trajectory feature instead of a transformer encoder.

\noindent\textbf{The Model without Appearance Feature (w/o App.)}. We input only trajectories with the spatial branch deleted, and consequently, the fused individual features are obtained by only concatenating the features from the temporal branch.

Table \ref{tab:abalation}  shows quantitative results compared with three alternative models on PANDA and JRDB-Group benchmarks, respectively.  The full model outperforms all the alternatives on both large-scale PANDA benchmark and small-scale JRDB-Group benchmark, which verifies the effectiveness of different modules. 
The model with only trajectory as input works better than the model with appearance feature but without occlusion encoder. The possible reason is that frequent occlusions make the appearance features full of noise, leading to a negative effect on performance.

To give a more intuitive reasoning, Figure  \ref{ablation} shows the performance of different ablation models on several typical cases.
We zoom in the people in the large-scale scenes, and the people belonging to the same group are marked with the same color. Our full model predicts the groups accurately. In the top row, the model without occlusion encoder and the model without transformer fail to distinguish the two separate people. The overlap of the two bounding boxes makes their appearance features similar, and therefore the edge between them is misclassified as positive. We also notice that the two people can be separated simply by the trajectory, demonstrating that improper use of appearance features or insufficient fusion model may cause confusion. In the middle row, four people follow the same trajectory along the sidewalk, and thus the model without appearance feature tends to detect the four people as a group. Appearance features are required to make the correct prediction. In the bottom row, six people are wrongly detected as a group by the model without the transformer. They have the same destination and similar posture, which will lead to the wrong detection if the features are insufficiently extracted.

\noindent

\noindent
\subsection{Robustness}
To evaluate the robustness of our method with a raw video input, we simulate the detection errors by adding noise to the bounding boxes or randomly dropping some detections on PANDA test set.

\noindent
\textbf{Adding Noise to Bounding Boxes.} Denote $\sigma$ as the noise intensity and $\textbf{u}=[x_0,y_0,x_1,y_1]$ as a bounding box, where $x_0$, $y_0$ and $x_1$, $y_1$ are the top-left and bottom-right coordinates of the bounding box. We define $w$ and $h$ as the width and the height of $\textbf{u}$.
Taking top-left coordinates as an example, we add the disturbances $\delta_x$ and $\delta_y$ to top-left coordinates $x_0$ and $y_0$ by sampling them from %norm 
Gaussian distributions with standard deviation $\sigma \times w$ and $\sigma \times h$, respectively. Table \ref{gs} shows the quantitative results with different levels of Gaussian noises.
Even with noise $\sigma=0.3$, our method still outperforms LSGD \cite{wang2020panda} that uses the clean input (Table \ref{cmpPANDA}).

\noindent
\textbf{Missing Detections.}
Denote \emph{MDR} as the missing detection rate of the bounding box to each person in the whole video. Table \ref{mdr} shows the quantitative results with different missing detection rates. With increasing \emph{MDR}, although the performance of our method degrades, our method still outperforms LSGD without missing detections \cite{wang2020panda},

\subsection{Limitations}
\begin{figure}[htbp]
    \centering
    \includegraphics[width=0.95\linewidth]{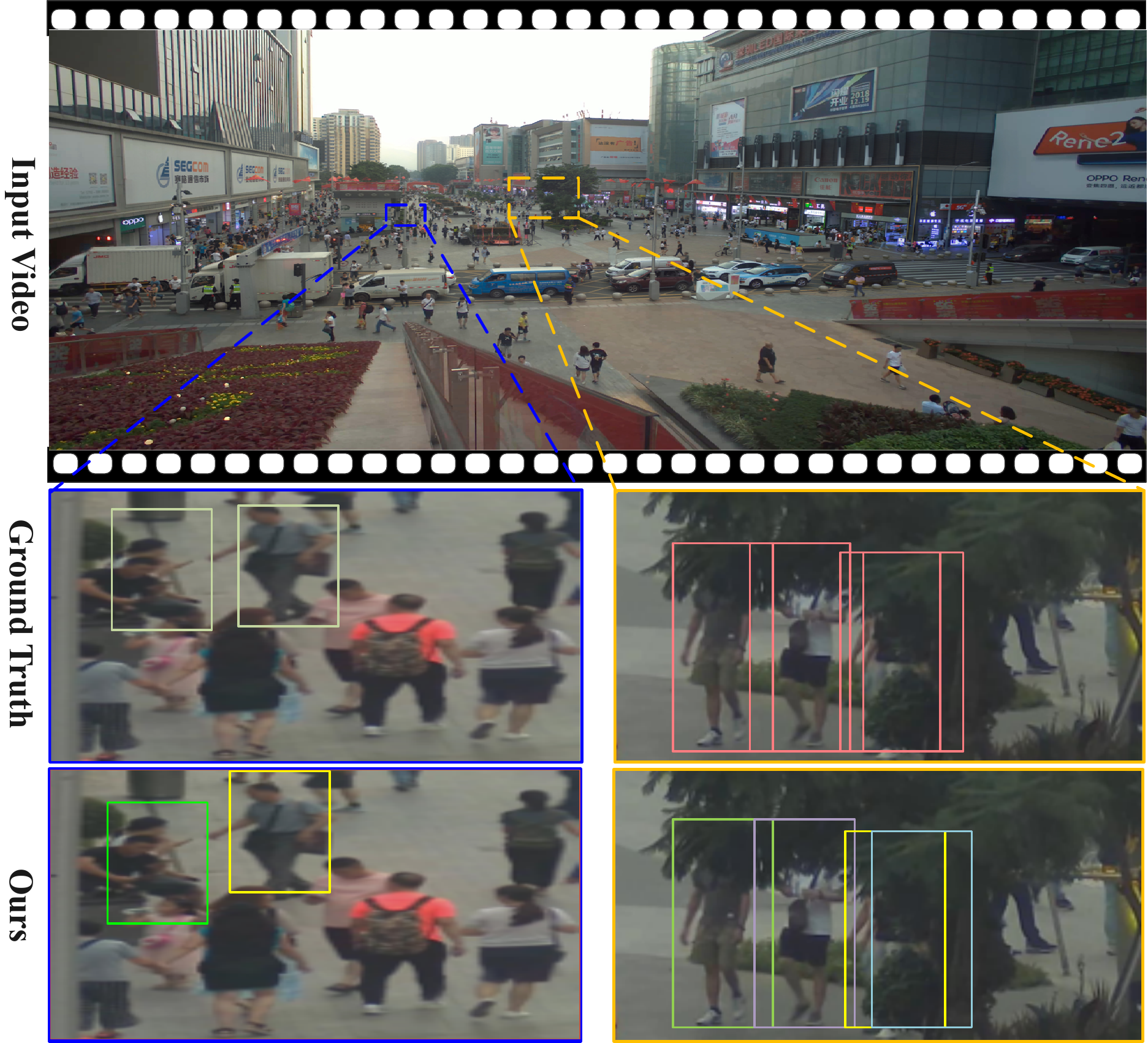}
    \caption{Some failure cases on large-scale scenes.}
    \label{PDfs}
\end{figure}

Although our method effectively detects groups in most large-scale scenes and small-scale scenes, it cannot cope with the cases of extremely low resolution and long-time occlusion. 
Figure \ref{PDfs} shows some failure cases on large-scale scenes. The middle row shows ground-truth groups and the bottom row shows our grouping results. Our method fails to group people due to the low resolution of images and the serve occlusion of people in a long time. 
This can be coped in our future work.

\section{Conclusion}
\label{con}
In this paper, we propose a novel end-to-end spatio-temporal framework for group detection in large-scale video scenes. The proposed framework addresses the challenges of frequent occlusions and complex spatio-temporal interactions. The occlusion encoder effectively extracts individual features by considering occlusion patterns, while the spatio-temporal transformers  capture the dynamic relationships among individuals in a hierarchical manner.
The comprehensive evaluation on the PANDA and JRDB-Group benchmarks confirms the superior performance of our method, and the ablation study demonstrates the effectiveness of each proposed module. 
The proposed framework opens up possibilities for further research in group detection and understanding. Future work could focus on exploring more advanced occlusion modeling and extending the framework to handle more complex scenarios with variable group sizes and activities.

% \newpage

\bibliographystyle{IEEEtran}
\bibliography{main}

\end{document}